%% file: paper.tex
\documentclass[sigconf, anonymous=False]{acmart}
\renewcommand\footnotetextcopyrightpermission[1]{}

\AtBeginDocument{%
  \providecommand\BibTeX{{%
    \normalfont B\kern-0.5em{\scshape i\kern-0.25em b}\kern-0.8em\TeX}}}

\acmConference[]{}{}{}

\usepackage{listings}
\usepackage{xspace}
\usepackage{subfig}
\usepackage[export]{adjustbox}
\usepackage{todonotes}
\usepackage{tablefootnote}

\usepackage{booktabs}
\usepackage{multirow}
\usepackage{threeparttable}

\newcommand\revision[1]{#1}

\definecolor{dkgreen}{RGB}{0,64,0}
\definecolor{ltgray}{RGB}{245,245,245}
\definecolor{mauve}{RGB}{139,0,139}

\begin{document}

\title{A Survey and Empirical Evaluation of Parallel Deep Learning Frameworks}

\author{Daniel Nichols, Siddharth Singh, Shu-Huai Lin, Abhinav Bhatele}
\affiliation{%
  ~\\
  \institution{$^{\dagger}$Department of Computer Science, University of Maryland}
  \city{College Park}
  \country{USA}
}
\email{{dnicho, ssingh37, slin185}@umd.edu, bhatele@cs.umd.edu}

\renewcommand{\shortauthors}{Nichols et al.}

\begin{abstract}
\input{abstract}
\end{abstract}



\maketitle

\section{Introduction}
\input{intro}

\section{Background}
\label{sec:bg}
\input{bg}

\section{Literature Survey}
\label{sec:survey}
\input{survey}

\section{Experimental Setup}
\label{sec:setup}
\input{setup}

\section{Comparative Evaluation}
\input{results}

\section{Conclusion}
\input{conc}


\bibliographystyle{ACM-Reference-Format}
\bibliography{cite,pssg}

\balance

\end{document}

%% file: abstract.tex
The field of deep learning has witnessed a remarkable shift towards extremely
compute- and memory-intensive neural networks. These newer larger models have
enabled researchers to advance state-of-the-art tools across a variety of
fields. This phenomenon has spurred the development of algorithms for
distributed training of neural networks over a larger number of hardware
accelerators. In this paper, we discuss and compare current state-of-the-art
frameworks for large scale distributed deep learning. First, we survey current
practices in distributed learning and identify the different types of
parallelism used. Then, we present empirical results comparing their
performance on large image and language training tasks. Additionally, we
address their statistical efficiency and memory consumption behavior. Based on
our results, we discuss algorithmic and implementation portions of each
framework which hinder performance.

%% file: intro.tex
The previous decade witnessed an explosion in the development of machine
learning algorithms. In particular, deep learning (DL), a subset of machine
learning focused on using neural networks for function approximation, has
gained widespread popularity. Deep neural networks (DNNs) have enabled the
advancement of the state of the art in a plethora of research areas: ranging
from visual recognition~\cite {BiT-resnet-eccv, vgg16-iclr, tao2020hierarchical,
zhao2019object-tnnls, vijayanarasimhan2017sfmnet} and natural language
processing~\cite{bert, gpt-2, nmt, nikolentzos2019message-aaai} to computational
chemistry and computer systems~\cite{jain:sc2013, bhatele:ipdps2015,
islam:sc2016, yeom:pmbs2016, marathe:sc2017, thiagarajan:ics2018,
thiagarajan:ipdps2018, menon:ipdps2020}.  Their popularity stems from the DNN's
ability to automatically learn low-dimensional representations from
high-dimensional unstructured data such as images, text and audio. Given enough
data, the representations learned by these models are often superior to
handcrafted features designed by domain experts.

\begin{figure}[h]
      \includegraphics[width=0.45\textwidth]{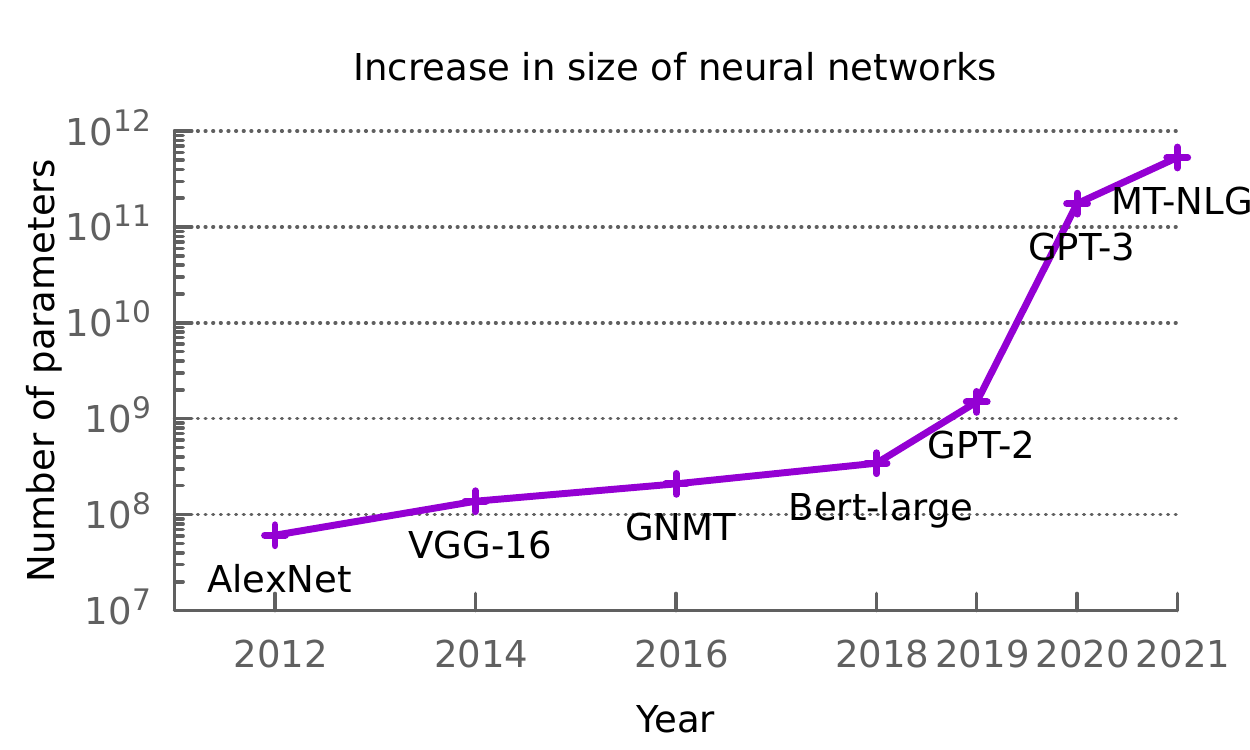}
    \caption{Neural networks have continued to grow in size in terms of the
number of parameters. Recent language networks have further contributed to this
trend.}
    \label{fig:model-sizes}
\end{figure}

The advances in accelerator technology, increased memory capacity per
accelerator, and faster networks have encouraged users of deep learning to
train neural networks with increasingly larger numbers of parameters. Figure
\ref{fig:model-sizes} shows the increasing number of parameters in the largest
networks since 2012. Often
times, it is impossible to train such networks on a single accelerator either due
to large execution time or insufficient memory capacity to fit these models.
The latter problem is further exacerbated for contemporary neural
architectures. For example, GPT-2, an extremely popular neural network used in
NLP requires 84 GB of GPU DRAM for training. This has motivated recent works in
parallelizing the task of deep learning: training large models using multiple
GPUs on a single node~\cite {huang2019gpipe_nips, kim2020torchgpipe} or across
multiple nodes connected by a network~\cite{sc2020zero, megatronlm, pytorchdist-vldb,
narayanan2019pipedream, you2018imagenet-icpp, jia2018data, dryden2019}.

Different parallel frameworks offer different strengths and weaknesses in
terms of performance (execution time for training), memory consumption, and
statistical efficiency. Ben-Nun et al.~\cite{bennun2019demystifying} surveyed parallel DL frameworks
and the different ways of exploiting the concurrency in neural networks in
2018. However, many new frameworks have emerged in the last three years, and
the authors limited their discussion to a qualitative analysis.  In this
paper, we survey the most popular parallel DL frameworks available today and
perform an empirical evaluation for the ones with open-source implementations
to compare various metrics. This comparative evaluation can help users of deep
learning select the best parallel framework for their training tasks.

We first present a comprehensive qualitative survey of the state of the art in
parallel deep learning. We classify approaches for parallelization into three
categories (defined in Section~\ref{sec:bg}): data parallelism, intra-layer
parallelism (sometimes referred to as model parallelism), and inter-layer
parallelism (sometimes referred to as pipelining,). We present the advantages
and disadvantages of using each approach and discuss the capabilities of
different frameworks that implement each type of parallelism.

An end user who needs a scalable DL framework for their training experiments
needs to know which frameworks provide the best statistical efficiency in the
shortest possible time. To the best of our knowledge, an empirical comparison
of parallel DL frameworks has not been attempted before. We identify two popular
training datasets and two neural networks to benchmark several open-source DL
frameworks including DDP~\cite{pytorchdist-vldb}, PipeDream~\cite
{narayanan2019pipedream}, ZeRO~\cite{sc2020zero}, Megatron~\cite{megatronlm}, 
TorchGPipe~\cite{kim2020torchgpipe}, and LBANN~\cite{essen2015lbann}.
We use metrics that matter the most to a deep learning researcher -- epoch  
execution times, statistical efficiency, and memory consumption. We run our 
experiments on two different supercomputers and clusters that are built using 
different generations of NVIDIA GPUs (A100s, V100s). Through these 
experiments, we seek to develop a consensus on the suitability of parallel 
frameworks to different scenarios.

In this paper we contribute:
\begin{itemize}
  \item A comprehensive survey of current state-of-the art techniques in 
        distributed deep learning organized by parallelization strategy.
  \item An empirical evaluation of these techniques across vision and language
        tasks on 2 different clusters that, to our knowledge, has not been 
        done before.
  \item A comparison of metrics, recorded across frameworks and architectures,
        that concern both the HPC and deep learning communities: runtime, 
        scaling, statistical efficiency, and memory consumption.
\end{itemize}

%% file: bg.tex
In this section, we first give brief descriptions of deep learning terminology. 
We refer the reader to \cite{Goodfellow-et-al-2016} for an in-depth review of deep learning.
We then provide an outline of the three ways in which training of a deep neural network 
can be parallelized: data parallelism, intra-layer parallelism and inter-layer parallelism.

\subsection{Definitions}


\noindent{\bf Neural networks:}
\revision{
Neural networks are parameterized functions for predicting properties of some
input data. They excel at learning low dimensional representations of complex,
high dimensional data.
}

\noindent{\bf Layers:}
\revision{
Networks are composed of a sequence of layers, which take the previous layer's 
output as input and computes some non-linear transformation.
}

\noindent{\bf Training and Loss:}
\revision{
The processing of finding the best parameters for a neural network is called
training. This is done by minimizing a loss function over an input data set.
Loss functions, such as mean squared error, are typically chosen to represent
the prediction capability of the network. 
}

\noindent{\bf Backpropagation:}
\revision{
Backpropagation is a dynamic programming algorithm based on reverse-mode 
automatic differentiation that computes the gradients of each layer with 
respect to the loss function.
}

\noindent{\bf Gradient Descent and Learning Rate:}
\revision{
Many training algorithms use variations of gradient descent to minimize the 
loss function. Gradient descent iteratively updates the parameters of the neural
network based on the negative gradient such that the loss moves towards 
a minima. The distance moved in the direction of the negative gradient is scaled
by a value called the learning rate.
}

\noindent{\bf Mini-Batches, Epochs and Stochastic Gradient Descent:}
\revision{
Computing gradients of the entire data set is expensive, so approximate
gradients are computed using random mini-batches of data.
This version of gradient descent is called batched stochastic gradient descent.
Each time the entirety of the data set is iterated over is called an epoch.
}

\noindent{\bf Statistical Efficiency:}
\revision{
Statistical efficiency is a measure of the relationship between epochs and 
accuracy/loss.
}
A training algorithm is said to be statistically efficient if it requires a low 
number of epochs to converge to a target validation loss. 

\subsection{Parallel Deep Learning Methods}

\begin{figure}[h]
    \centering
      \includegraphics[width=2.8in]{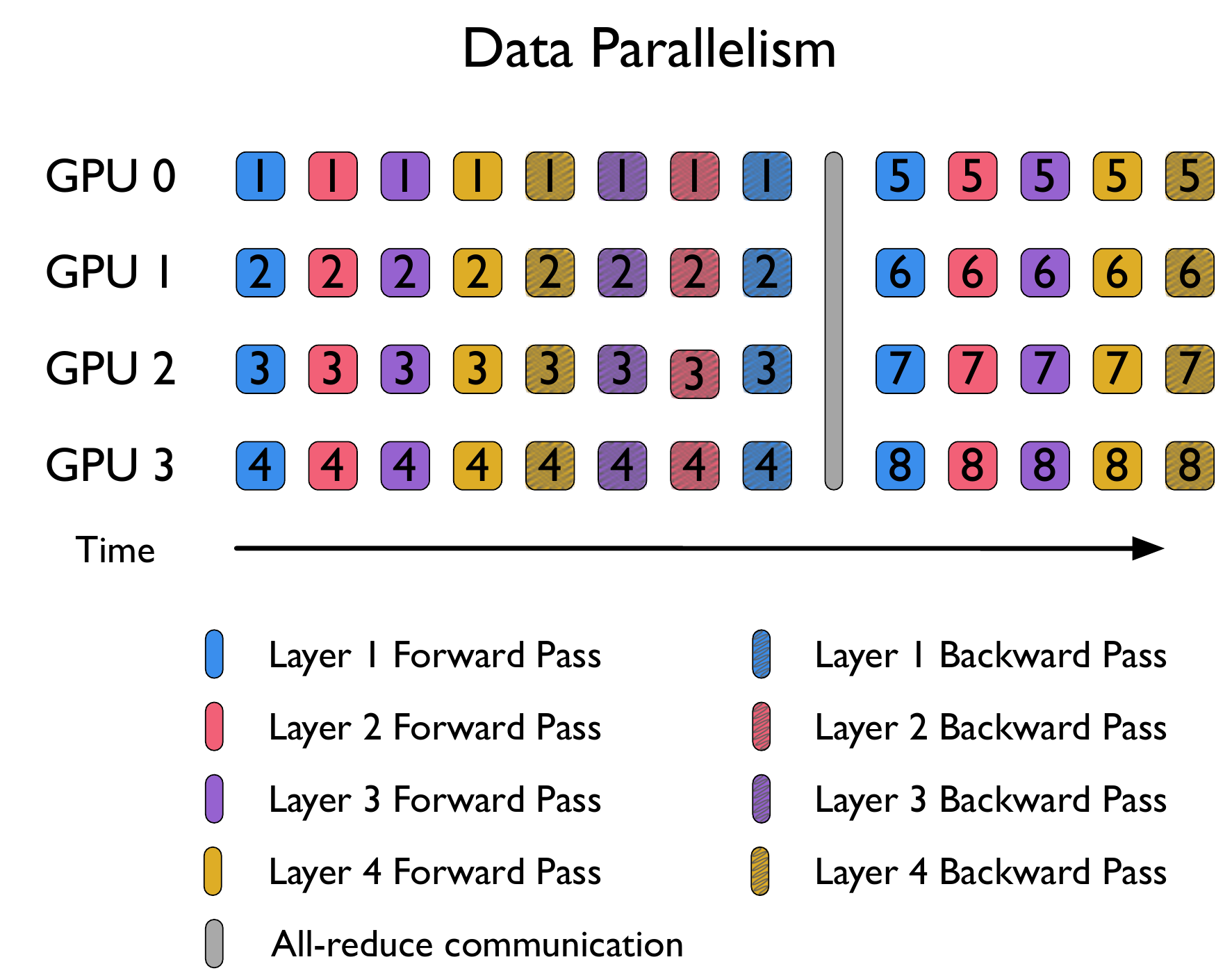}
    \caption{Processing of mini-batches over time in data parallelism. 
      Each GPU has a copy of all the layers (shown in different colors) and different mini-batches (numbered) are processed by different GPUs.}
    \label{fig:data-parallel}
\end{figure}

\noindent{\bf Data Parallelism:}
Data parallelism refers to an even division of training data among worker GPUs. 
Each GPU possesses a copy of the neural network along with it's parameters. 
Gradient calculation via backpropagation proceeds independently on all GPUs. 
These gradients are then subject to a collective all-reduce operation before 
the weight update step of the optimizer. 
The all-reduce step can either take place synchronously after each 
mini-batch, or asynchronously using a central parameter server. 
Implementations of data parallelism are widely 
available in popular deep learning frameworks like PyTorch~\cite{pytorchdist-vldb}, 
and TensorFlow~\cite{tensorflowosdi2016}.
Figure \ref{fig:data-parallel} illustrates data parallelism across 4 GPUs. 


\noindent{\bf Intra-layer Parallelism:}
Intra-layer parallelism distributes the work of a layer by dividing its 
computation across multiple GPUs. Parallelizing an entire neural network 
entails applying intra-layer parallelism to some or all of its constituent 
layers. Research in this area is focused on optimizing the 
multi-GPU execution of different kinds of layers - Fully Connected, 
Convolutional~\cite{pmlr-v28-coates13, oyama2020case, mesh_tf} and more 
recently the Transformer~\cite{megatronlm}. Intra-layer parallelism enables us to train neural 
networks that would not fit inside the DRAM of a single GPU. 

\begin{figure}[h]
    \centering
      \includegraphics[width=\columnwidth]{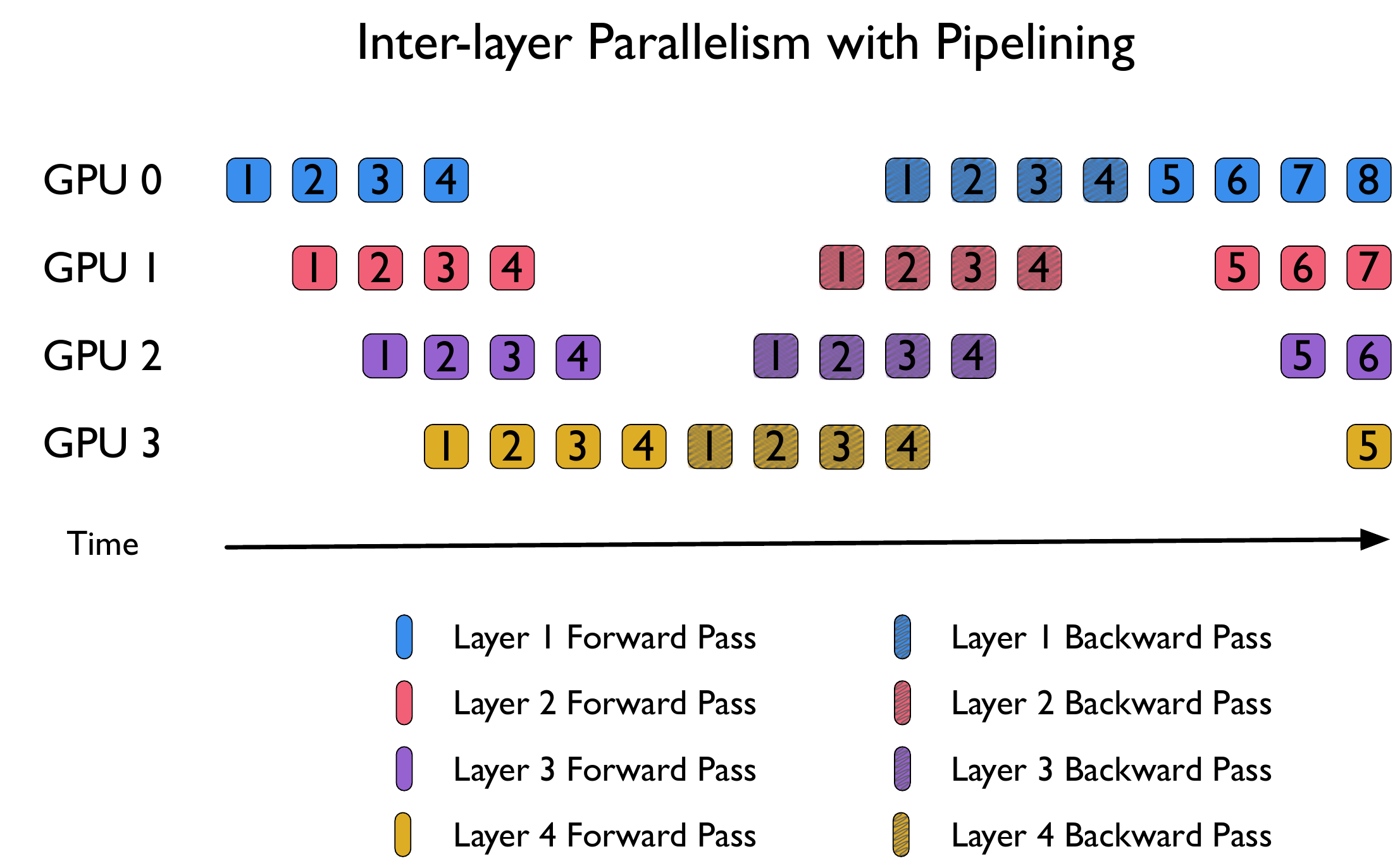}
    \caption{Processing of micro-batches in inter-layer parallelism. 
      Each GPU holds one or more layers in the network and all mini-batches pass through all the layers/GPUs.}
    \label{fig:model-parallel}
\end{figure}

\noindent{\bf Inter-layer Parallelism:}
In inter-layer parallelism contiguous subsets of layers are mapped to 
individual GPUs. Each GPU is thus tasked with operating on a subset of the neural 
network. Exchange of activations and gradients 
among consecutive layers on different GPUs takes place via point-to-point 
communication primitives. To achieve true parallelism more
than one mini-batch should be active on different GPUs at a time since the
processing of a mini-batch across layers is sequential and cannot be
parallelized. This is called pipelining. The maximum number of mini-batches active in
the system at any given point of time is called the pipeline limit. Figure
\ref{fig:model-parallel} shows inter-layer parallelism in
action with four GPUs and a pipeline limit of four. Just like intra-layer parallelism 
inter-layer parallelism makes it possible to train models whose memory requirements 
exceed the DRAM capacity of a single GPU.

\subsection{Related Work}
\label{sec:related-work}

Pouyanfar et al.~\cite{pouyanfar2018dlsurvey} and Ben-Nun et al.~\cite{bennun2019demystifying} comprehensively survey established techniques in sequential deep learning as well as distributed.
Another survey \cite{sze2017efficientdnn} covers work in processing neural networks efficiently.
Distributed training on big data software stacks (such as Spark and Hadoop) is explored by Lu et al.~\cite{lu2018dlbigdata}.
The network demands of parallel training are presented in \cite{awan2020dlcommunication} where typical communication workloads are profiled and characterized.
Tang et al.~\cite{tang2020communicationefficient} further character distributed training communication via analytical models and survey current practices.
\revision{
  We also point the reader to the MLPerf benchmarks\footnote{\url{https://mlcommons.org/en/training-normal-07/}}, which have become popular
  for comparing deep learning algorithms, frameworks, and hardware. 
}


%% file: survey.tex
In this section we present a survey of current state-of-the-art techniques and implementations for
each type of distributed learning. Table \ref{tab:survey-summary} provides an overview
of each discussed framework.

\input{tables/survey-summary}


\subsection{Data Parallelism}\label{sec:data-parallelism}
\input{survey-data}

\subsection{Intra-Layer Parallelism}\label{sec:intra-layer-parallelism}
\input{survey-intra}

\subsection{Inter-Layer Parallelism}\label{sec:inter-layer-parallelism}
\input{survey-inter}

%% file: tables/survey-summary.tex
\begin{table*}[t]
\begin{threeparttable}
\caption{Summary of Literature Review on Parallel Deep Learning}
\label{tab:survey-summary}
\begin{tabular}{@{}llrr@{}}
\toprule
Framework      & Type of Parallelism & \begin{tabular}[c]{@{}l@{}}Largest\\Accelerator Count\end{tabular} & \begin{tabular}[c]{@{}l@{}}Largest Trained Network\\(No.~of Parameters)\end{tabular} \\ \midrule
FlexFlow       & Hybrid              & 64 GPUs                & 24M$^{*}$              \\
PipeDream$^{**}$        & Inter-Layer         & 16 GPUs                 & 138M                  \\ 
DDP$^{**}$            & Data                & 256 GPUs               & 345M             \\
GPipe          & Inter-Layer         & 8 GPUs                 & 557M
\\
MeshTensorFlow & Intra-Layer         & 512-core TPUv2         & 4.9B                  \\  
Megatron$^{**}$       & Intra-Layer         & 512 GPUs               & 8.3B                  \\
TorchGPipe$^{**}$     & Inter-Layer         & 8 GPUs                 & 15.8B                  \\ 
KARMA          & Data                & 2048 GPUs              & 17B              \\  
LBANN$^{**}$          & Data                & 3072 CPUs              & 78.6B                  \\
ZeRO$^{**}$           & Data                & 400 GPUs               & 100B            \\
ZeRO-Infinity  & Data & 512 GPUs & 32T \\ 
AxoNN  & Inter-Layer & 384 GPUs & 100B \\
 \bottomrule
\end{tabular}
\begin{tablenotes}
  \small 
  \item $^{*}$Note: FlexFlow does not provide a parameter size for the largest network it trains. We have defaulted to the
  largest network with a known network size cited in their paper.
  \item \revision{ $^{**}$The following frameworks are compared quantitatively in Section \ref{sec:setup}}
\end{tablenotes}
\end{threeparttable}
\end{table*}

%% file: survey-data.tex

Data parallelism has been the go-to algorithm for parallelizing neural network 
training. 
It is simple in design and performs well with the correct settings.

\subsubsection{Small Models}

Data parallelism hinges on a synchronous all-reduce operation to gather the 
gradients across all GPUs. 
Naturally, this can become a bottleneck as the size of the gradients being 
being shared grows.
This problem is further exacerbated by the increasing computational 
capabilities of hardware accelerators. The ensuing decrease in the computation 
to communication ratio increases the severity of this problem.

Initial attempts to reduce the communication 
overhead targeted introducing asynchrony in the stochastic gradient descent (SGD) 
algorithm \cite{niu2011hogwild-nips, projectadam, distbelief}. However, Chen et al. \cite{chen2016revisiting} demonstrate that synchronous SGD and its 
variants converged faster with higher accuracy than their asynchronous counterparts. 

Efforts to minimize communication bottlenecks continued. Zhang et al. \cite{wait-free-backprop-usenix}
devise a strategy known as Wait-Free Backpropagation (WFBP) to interleave GPU and CPU computation and 
communication. WFBP reduces bursts in network traffic and lowers overall network strain. 
Using WFBP, Zhang et al. achieve speed-ups in training times in 16 and 32 single-GPU machines.
WFBP has become the de-facto approach for data parallelism frameworks. 

PyTorch DistributedDataParallel (DDP)~\cite{pytorchdist-vldb}, Horovod~\cite{sergeev2018horovod} and Livermore Big Artificial Neural Network (LBANN)~\cite{essen2015lbann} toolkit are three 
open source frameworks designed to assist in transitioning models into a 
distributed environment. Out of these frameworks PyTorch DDP has been extremely 
popular among the deep learning community due to its seamless integration with 
PyTorch~\cite{paszke2019pytorch-nips}. Horovod is an implementation of WFBP for 
TensorFlow by Uber. LBANN accelerates parallelized deep learning by taking 
advantage of high performance computing hardware. These implementations share 
an uncanny similarity in 
the way they optimize WFBP. Instead of having an individual all-reduce call for 
each parameter tensor, they fuse parameter tensors into fixed size bins. All 
reduce calls are made at the granularity of these fused parameter bins. This 
increases network bandwidth utilization and thus the overall performance of 
these frameworks. Although the fused tensor bin-size is kept as a tunable 
hyperparameter, Li et al.~\cite{pytorchdist-vldb} demonstrate that the default 
bucket size of PyTorch DDP i.e. 25MB is a reasonable choice for efficient 
scaling. 

\subsubsection{Large Models} \label{dp_mem}
Given the abundance of large training datasets neural networks with 
increasingly larger number of parameters have led to 
tremendous gains in performance on a variety of training tasks. As models and 
datasets grow in size GPU memory capacity becomes a 
major bottleneck. Data parallelism requires each GPU to store its own copy of 
the neural network. With larger models and datasets the memory required to 
house the activations, gradients and parameters of these neural networks 
often exceeds the capacity of a single GPU DRAM. Data parallelism is thus 
rendered infeasible for training large models without memory optimizations.

Zero Redundancy Optimizer (ZeRO)~\cite{sc2020zero} is a framework built over 
PyTorch to reduce per-GPU memory consumption. 
The paper observes that most memory during training is occupied by optimizer 
states, gradients, and parameters. ZeRO partitions these model states across 
GPUs to remove memory redundancies. With ZeRO, memory reduction scales 
proportionally with the number of GPUs while communication 
overhead only increases by a constant factor of 1.5x. The paper finds 
improvements in model size, training performance, and scalability with 100 
billion parameter models on up to 400 GPUs using the Adam optimizer \cite{KingmaAdam2014} and mixed precision. Researchers at Microsoft have used ZeRO 
to train one of the largest neural networks in language modeling literature: a 
17B parameter neural network called the Turing-NLG. 

Out-of core training algorithms like NVIDIA's vDNN~\cite{vDNN} are often used 
to train neural networks on a single GPU 
with insufficient DRAM capacity. These algorithms move data back and forth 
between the CPU and the GPU to free up space on the GPU. 
KARMA~\cite{wahib2020scaling-sc} is a framework built over PyTorch that extends 
this out-of-core approach to data parallelism on multiple GPUs. They design an 
efficient algorithm for automatic offloading and prefetching of activations and 
parameters of the neural network to and from the CPU DRAM. 
These capabilities are further extended to support multi-GPU models by 
performing weight updates on the CPU. KARMA sees a 1.52x speed-up against 
other state-of-the-art out-of-core methods. It provides an efficient way to 
utilize data parallelism for large models that would otherwise necessitate 
other frameworks. Zero-Infinity \cite{zero_infinity} is another framework that provides support for out-of-core data parallel training for multi-billion parameter models. Using their memory optimizations, The authors are able to deploy a 32 trillion parameter model on as little as 512 GPUs while maintaining a decent throughput of around 40\% of the peak.

\subsubsection{Large Effective Mini-Batch Sizes} \label{lars}
Data parallelism is most efficient with high per-GPU workloads. This is ensured 
by fixing the per-GPU mini-batch size. As an example, suppose a ResNet model 
with a per-GPU mini-batch size of 128 is trained over 64 GPUs. This is 
equivalent to an effective mini-batch size of 8192 on a single 
GPU. \revision{It has been empirically shown that an extremely large effective mini-batch size has an adverse effect on the statistical efficiency of neural network training \cite{yanyou-large-batch}}.  

The naive approach to compensate for this is to increase the 
learning rate (LR). Krizhevsky \cite{alexnet} proposes to scale 
LR linearly with mini-batch size. Problems emerge as more workers are added to 
accelerate training: large LR values result in accuracy losses and training 
instability. 

Goyal et al. \cite{yanyou-large-batch} propose a LR warmup scheme 
to combat accuracy loss. Training begins with a lower LR that slowly builds up 
to a target value following the linear scaling rule. The paper was able to train 
ResNet-50 with a mini-batch size of 8K and accuracy matching smaller mini-batch models. 

You et al.\cite{you2017large, you2018imagenet-icpp} devise Layer-wise Adaptive Rate Scaling (LARS) as an alternate 
approach to LR warmup. LARS adapts the global LR to create separate LRs per model 
layer based on the ratio between layer weights and gradient updates. The paper 
observes this ratio varies across layers and provides insight into the efficacy of a 
layer’s weight updates. You et al. utilize LARS to train AlexNet and ResNet-50 with a 
mini-batch size of 32K without accuracy loss.

LARS experiences inconsistent performance gains across different deep learning tasks. 
You et. al \cite{you2019large} propose a general strategy to adapt any iterative optimizer for large mini-batch training. 
They apply this strategy to create LAMB using the Adam optimizer as a base. 
Using LAMB, You et al. scale BERT training to a mini-batch size of 32K without performance degradation.

%% file: survey-intra.tex

State of the art training techniques in intra-layer parallelism span from fine-grained parallel implementations of numerical kernels to dividing the coarse-grained work of a single layer across processes.
It is often used in conjunction with other parallelization strategies such as data or inter-layer parallelism.

\subsubsection{Fine-Grained Parallelism}\label{sec:intra-fine-grain}

At the fine-grained level many techniques draw from existing numerical methods and adapt them to deep learning.
Matrix multiplication and convolutions are the most utilized kernels and have been the focus of much optimization from the ML and broader scientific community.
Many accelerators and processors have paired software libraries which implement these kernels tuned to their hardware such as CuDNN\cite{cudnn}, MIOpen\cite{jeh2019miopen}, and OneDNN.

Accelerators have been at the core of fine-grained parallelism within a layer.
Several works have introduced techniques, some ML based, for mapping layer computations to the hardware optimally\cite{gamma-iccad2019, lift-cgo2017, maestro-micro2020}.
Here a mapping is the tiling strategy, computation order, and parallelization strategy, hence, the search space for optimal mappings can be immense.

There has been recent interest in using hardware accelerators other than GPGPUs to train deep networks.
FPGAs have emerged as a viable candidate in DNN acceleration due to their lower energy consumption than GPUs and the flexibility provided by their reconfigurability.
Recent work has explored optimizing DNN operations on FPGA hardware\cite{cnn-fpga-isfpg2017}.
More recently, novel architectures have been proposed to improve memory re-use and parallel performance\cite{neurocube-isca2016, eyeriss-isca2016, eyeriss-ssc2017}.

\subsubsection{Coarse-Grained Parallelism}\label{sec:intra-coarse-grain}

Orthogonal to the fine-grained compute kernels there have been techniques developed to divide work inside a layer along coarser tensor dimensions.
These typically involve using optimization algorithms and/or ML to identify optimal partitions of computation and data within a layer and then developing a parallel strategy for execution.
Song et al. propose a method for finding communication optimal parallel strategies on accelerator arrays in linear time\cite{hypar-hpca2019}.
Similarly, Jia et al. introduce a novel Markov Chain Monte Carlo based search for finding optimal parallelization strategies, which encompasses intra-layer in its operator dimension\cite{jia2018data}.

MeshTensorFlow accomplishes a similar effect by mapping tensor dimensions to a n-dimensional processor array or "mesh"\cite{mesh_tf}. 
\revision{
    These tensors are split and/or replicated across the mesh, such that the computation can be done in parallel using the processor array.
    The framework itself provides an interface for users to define a layout.
    Any layout will produce the same results for the same problem, however, the memory footprint and performance can be
    greatly improved with an optimal layout.
}

Dryden et al\cite{dryden2019channelparrallelism} also propose several algorithms for partitioning convolution tensor dimensions with the goal of reducing all-reduce time during training.
Their algorithms are available in the LBANN framework.
Convolutions are also parallelized in \cite{oyama2020case} with a hybrid parallelism by extending data parallelism with parallelism in the spatial domain.
For language-based models Megatron\cite{megatronlm} achieves a similar parallelism by partitioning the blocks in transformer layers across processors.
Megatron has been increasingly used as language models become more common and larger (see Figure \ref{fig:model-sizes}).
It has shown up to 74\% weak scaling coefficient on 512 GPUs.

Dividing layer tensor dimensions across processors is, however, very sensitive to the layer type.
For instance, fully connected layers involve an all-to-all computation and therefore all-to-all communication, which is more expensive the data parallelism's allreduce.
Thus, it is hard to generalize coarser grained intra-layer parallelism for models with custom layers.
To combat this some methods look strictly at compute graph operations and not model layers \cite{jia2018data}.

%% file: survey-inter.tex


True inter-layer parallelism can only be achieved by pipelining i.e. having 
multiple mini-batches active in the system at any given instance. There are two ways 
to achieve pipelining: with and without flushing. In this section, we discuss 
the pros and cons of both approaches. We also provide an overview of frameworks 
that implement these approaches. 

\subsubsection{Pipelining with Flushing}
Pipelining with flushing divides a mini-batch into micro-batches of equal size. These 
micro-batches are injected one by one into the system. GPUs accumulate gradients 
from all the micro-batches in the system. A GPU updates its weights only after 
it has finished the backward pass of the last micro-batch. The next mini-batch and 
its corresponding micro-batches are injected after all the 
GPUs have finished updating their weights. This approach to pipelining is also called 
micro-batching. \revision{The number of micro-batches is usually kept to be much larger 
than the number of workers so that each worker can compute concurrently.} Ensuring optimum hardware 
utilization requires having a large mini-batch size. To maintain statistical 
efficiency at large mini-batch sizes the same set of solutions discussed in Section 
\ref{lars} can be used. \revision{Figure \ref{fig:model-parallel} shows pipelining with flushing in action. Worker GPUs incur idle time between the forward pass of the last micro-batch and the backward pass of the first micro-batch. These are called pipeline bubbles. They reduce the overall hardware utilization of the system}  A load balanced mapping of layers to GPUs is 
absolutely critical to maximize performance. The load balancing algorithm must 
also be communication-aware. This is because activations and gradients 
exchanged at GPU boundaries can be in the magnitudes of GBs for large neural 
networks. An efficient implementation of pipelining with flushing must have 
load balancing support.    

\input{tables/systems}

This idea was first introduced by Huang et al. in GPipe~\cite{huang2019gpipe_nips}. 
Using GPipe they trained a 557M parameter neural network - AmoebaNet-B~\cite{amoebanet-aaai} on the ImageNet~\cite{ILSVRC15} dataset and surpassed the state of 
the art in a number of downstream image classification tasks. TorchGPipe~\cite{kim2020torchgpipe} is an unofficial open-source implementation of GPipe built 
on the PyTorch~\cite{paszke2019pytorch-nips} backend. GEMS (GPU-Enabled Memory Aware 
Model-Parallelism System)~\cite{GEMS} introduces a novel approach to increase 
hardware utilization. This framework proposes an algorithm to train two neural 
networks concurrently using pipelining without flushing on multiple GPUs. They 
double the throughput of the system by overlapping the forward and backward 
passes of the two neural networks. We refer the reader to their paper for the 
details of their implementation. Recently ZeRO~\cite{sc2020zero} and 
Megatron~\cite{megatronlm} also extended support for this approach towards 
inter-layer parallelism. TorchGPipe~\cite{kim2020torchgpipe} provides a load balancing algorithm that 
seeks to balance the net execution time of the forward and backward pass of a 
micro-batch on each GPU. However, their algorithm ignores the communication 
overhead of exchanging tensors across GPU boundaries. Megatron divides the 
layers of a transformer across GPUs, which is optimal because all the 
layers of a transformer are identical. ZeRO also provides an identical strategy 
that divides the layers equally across GPUs. Additionally, they also support a 
load balancing algorithm that equalizes GPU memory consumption across GPUs. 
AxoNN \cite{singh:ipdps2022} introduced a novel asynchronous communication 
backend for inter-layer parallelism. To the best of our knowledge this is the 
first work that utilizes asychrony for increasing hardware utilization by 
opting for MPI instead of NCCL. They also introduce a memory optimization 
algorithm that they use to decrease the pipeline depth, increase data 
parallelism and outperform the state-of-art by 15\%-25\% on models with as many 
as 100 billion parameters.
 
\subsubsection{Pipelining without Flushing}
In this approach the number of mini-batches active in the system is kept constant. 
As soon as a mini-batch finishes its backward pass on the first GPU a new mini-batch is 
injected into the system to maintain full pipeline occupancy. Unlike pipelining 
with flushing, weight updates on a GPU take place as soon as it is done with 
the backward pass of a mini-batch. This method of pipelining seeks to increase 
hardware utilization by removing flushing induced bubbles in the pipeline. 
However, statistical efficiency of such a training algorithm reduces drastically. 
This is due to a problem called weight staleness that occurs when newer mini-batches in a pipeline 
encounter stale weights in forward passes which are yet to be updated with the 
backward pass of older mini-batches. This is one of the major reasons why pipelining 
without flushing has not seen widespread adoption. PipeDream~\cite{narayanan2019pipedream} is a framework that implements pipelining without 
flushing. \revision{It employs an algorithm called weight stashing to counter weight staleness. We refer the reader to their paper for exact details of the implementation.}
Chen et al.~\cite{chen2019efficient} suggest 
predicting future weights from stale weights using a variant of SGD with 
momentum~\cite{SGDMomentum}. PipeDream additionally proposes a static load balancing 
algorithm that is communication aware. It instruments each layer and uses the 
profiling data in its load balancer. Their framework also has an 
additional provision to replicate compute-intensive layers across GPUs to 
increase their throughput. Replicated layers synchronize their gradients via 
all-reduce after each backward pass. 

\input{tables/datasets}

%% file: tables/systems.tex
\begin{table*}[t]
\begin{threeparttable}
\caption{System information about the HPC platforms used for the experiments.}
\label{tab:system-info}
\begin{tabular}{@{}lrcrcrrrr@{}}
\toprule
\textbf{System} & \textbf{No. of Nodes} & \textbf{CPU} & \textbf{\begin{tabular}[c]{@{}l@{}}Cores/\\node\end{tabular}} & \textbf{GPU} & \textbf{\begin{tabular}[c]{@{}l@{}}GPUs/\\node\end{tabular}} & \textbf{\begin{tabular}[c]{@{}l@{}}CPU Mem. /\\ Node (GB)\end{tabular}} & \textbf{\begin{tabular}[c]{@{}l@{}}GPU Mem. /\\ Node (GB)\end{tabular}} & \textbf{\begin{tabular}[c]{@{}l@{}}GPU FP64 \\ Peak (TFlop/s) \end{tabular}} \\ \midrule
Lassen          & 795                   & IBM Power9 &  44 &  NVIDIA V100 & 4       & 256                                                 & 64                                                                   & 7.0 \\
ThetaGPU       & 24                    & AMD Rome & 64     & NVIDIA A100  & 8        & 1024                                               & 320                                                                  & 9.7 \\
\end{tabular}
\centering
\end{threeparttable}
\end{table*}

%% file: tables/datasets.tex
\begin{table*}[t]
    \caption{Training datasets and network hyperparameters used for benchmarking in the paper}
    \label{tab:datasets}
    \begin{tabular}{@{}llllrcrrr@{}}
    \toprule
    \textbf{Dataset} & \textbf{\begin{tabular}[c]{@{}l@{}}Training \\Split Size\end{tabular}} & \textbf{\begin{tabular}[c]{@{}l@{}}Validation \\Split Size\end{tabular}} & \textbf{Network} & \textbf{\begin{tabular}[c]{@{}l@{}}Mini-Batch Size \\per GPU\end{tabular}}  & \textbf{\begin{tabular}[c]{@{}l@{}}Optimizer$^{\dagger\dagger}$ \end{tabular} } &  \textbf{\begin{tabular}[c]{@{}l@{}} Learning \\ Rate \end{tabular}} & \textbf{\begin{tabular}[c]{@{}l@{}} No. of \\ Epochs \end{tabular}} & \textbf{\begin{tabular}[c]{@{}l@{}} L2 Decay \end{tabular}}   \\ \midrule
    ImageNet        &  1,281,167          & 50,000                & VGG-16           & 64$^{\dagger}$  & SGD$^{\dagger}$ & 0.01$^{\dagger}$  & 90$^{\dagger}$  & 0.0001$^{\dagger}$  \\ \midrule
    Wikitext-103    & 103,227,021         & 217,646               & GPT2-medium       & 32$^{**}$       & LAMB$^{*}$     & 0.001$^{*}$  & 100$^{**}$ & 0.01$^{*}$  \\ \bottomrule
    \end{tabular}
    \begin{tablenotes}
        \small 
        \item $^{*}$ Values directly taken from MLPerf
        \item $^{**}$ Values defined as unconstrained in MLPerf
        \item $^{\dagger}$ Values directly taken from torchvision - \url{https://github.com/pytorch/vision/tree/master/references/classification} 
        \item $^{\dagger\dagger}$ For ZeRO, we use the Adam optimizer with 0.001 learning rate and 0.01 l2 decay as it's memory optimizations only work with Adam
      \end{tablenotes}
\end{table*}

%% file: setup.tex
In this section we present a detailed overview of our empirical evaluation of 
a number of parallel deep learning frameworks.  
\subsection{Choice of Frameworks}\label{sec:choice-of-frameworks}


We use 
DDP\footnote{\url{https://github.com/pytorch/pytorch}\quad @1.8.0}~\cite{pytorchdist-vldb}, 
ZeRO\footnote{\url{https://github.com/microsoft/DeepSpeed}\quad @0.3.13}~\cite{sc2020zero}, 
Megatron\footnote{\url{https://github.com/NVIDIA/Megatron-LM}\quad @2.3}~\cite{megatronlm}, 
PipeDream\footnote{\url{https://github.com/siddharth9820/pipedream}\quad @00931df}~\cite{narayanan2019pipedream}, 
TorchGPipe\footnote{\url{https://github.com/kakaobrain/torchgpipe}\quad @a1b4ee2}~\cite{kim2020torchgpipe}, 
LBANN\footnote{\url{https://github.com/LLNL/lbann}\quad @0.101}~\cite{essen2015lbann}, and
AxoNN\footnote{https://github.com/hpcgroup/axonn/\quad @db1c6a0}~\cite{singh:ipdps2022}
for our empirical analysis. For Megatron we profile it's implementations of 
data-parallelism and intra-layer parallel implementations separately. We refer to these as Megatron-data and Megatron-intra respectively. 
This subset is representative of the three types of parallelism discussed in 
Section \ref{sec:survey}. We select frameworks which have open-source 
implementations, are easy to setup, and have a relatively large 
user-base. We also tried to include MeshTensorFlow~\cite{mesh_tf} and FlexFlow 
~\cite{jia2018data} in our set of frameworks. However, despite our best 
efforts we could not set them up successfully for experimentation on our 
machines.

To prevent dataloading from being a bottleneck we copy training  
datasets into node-local SSDs before training. Data is loaded using PyTorch's 
distributed data loader with several worker processes. We defaulted to four 
processes, separate from the main process, to read in data. MegatronLM 
implements their own data loaders, which we used with Megatron rather than 
PyTorch's. In practice we found these to be much faster than the default 
PyTorch data loaders.

For a fair performance evaluation of each framework we used mixed 
precision on the V100 and A100 cards on Lassen and 
ThetaGPU \cite{micikevicius2018mixed}. Of the frameworks we ran DDP, Megatron, 
LBANN, and ZeRO were the only ones that supported mixed precision with 
distributed training. 

All of the listed frameworks use Pytorch 1.8.0, CUDA 11.0, and CuDNN 8.0 for 
launching computation on GPUs. For inter-GPU communication, PipeDream uses the 
gloo communication library shipped with Pytorch 1.8.0, whereas all of the other 
frameworks use NCCL 2.7.8.

\subsection{System Hardware}

Table \ref{tab:system-info} describes the systems and hardware used in our 
training. Lassen is an IBM machine at Lawrence Livermore National Laboratory with a 
Mellanox network. 
It currently sits at number 26 on the Top500 list. 
ThetaGPU is a GPU extension of the Cray XC40 Theta system.

Each system was selected to be representative of typical machines used for DL 
training.
Lassen is similar to other leadership HPC systems with GPU-dense nodes.
The ThetaGPU extension of Theta with dense A100 nodes is more typical of 
current cutting edge AI machines. 

\subsection{Datasets and Neural Networks}

We evaluate the aforementioned subset of frameworks on two popular deep 
learning tasks: image classification and language modeling. For the former 
task we use The ImageNet Large Scale Visual Recognition Challenge (ILSVRC) 
2012 dataset~\cite{ILSVRC15}. This dataset has been widely used to train large 
state of the art image classification neural networks throughout the last 
decade. It consists of more than a million RGB images of dimension 224x224 
evenly divided across 1000 image classes. We use this dataset to train the 
VGG-16~\cite{vgg16-iclr} architecture on our selected 
subset of frameworks. Language modeling is an unsupervised learning task 
wherein models are trained to predict the next word in a sentence given all of 
the previously occurring words. We use the Wikitext-103~\cite{wikitext-103} dataset for our 
language modeling training workloads. This dataset is comprised of more than 
28000 articles from the English Wikipedia amounting to a total of 100 million 
English words. Language modeling has gained immense popularity recently in NLP 
for training extremely large neural networks. Researchers have achieved stellar 
performance with these models in a variety of downstream tasks like question 
answering, textual entailment, translation, reading comprehension, etc... We 
train the GPT-2-medium architecture proposed by OpenAI in their paper \cite
{gpt-2} on the Wikitext-103 \cite{wikitext-103} 
dataset. Table \ref{tab:datasets} provides an overview of the datasets
used across our experiments. 

\subsubsection{Hyperparameters}

The epoch execution times and statistical efficiency of a training algorithm 
are very sensitive to the choice of hyperparameters. Learning rate schedules, 
optimizer choices and weight decay values can have a large impact on the 
statistical efficiency. Larger mini-batch sizes reduce epoch execution times 
at the expense of statistical efficiency. 


\revision{
    Hyperparameters were chosen based on corresponding MLPerf~\cite{MLPerf2020} benchmarks,
    which are a standard means of comparison for DL training.
    Because of this we keep the parameters fixed between frameworks.
    For parameters not included in the MLPerf description we choose them 
    based on the values given in their respective papers.
    We ensure that training with our hyperparameters gives us reasonable performance 
    on the validation set. Table \ref{tab:datasets} provides an overview of the 
    hyperparameters applied to each model. It is possible further tuning
    could improve the performance and/or statistical efficiencies.
}

For efficient scaling to larger GPU counts, data parallel algorithms 
typically use a fixed mini-batch size per GPU to maintain a constant computational workload per GPU. 
Thus, to ensure a fair comparison of other frameworks with DDP, AxoNN, 
ZeRO, LBANN and Megatron-data we do the following for each framework:
\begin{itemize}
    \item Megatron-intra - We linearly scale the mini-batch size with increasing number of GPUs.
    \item TorchGPipe - We fix the size of a micro-batch and set the number of micro-batches to 4 times that of the GPU count. 
    \item PipeDream -  We fix the size of a mini-batch. PipeDream ensures constant computational workload on each GPU by increasing it's pipeline limit automatically.  
\end{itemize}

\subsection{Exceptions}\label{sec:exceptions}

We make the following exceptions to the experimental setups listed above. We only 
show results for PipeDream on a subset of the GPUs due to the framework deadlocking on higher
GPU counts.
We only show results for TorchGPipe upto 8 GPUs on ThetaGPU and 4 GPUs on 
Lassen as it is only applicable to a single node. We only show results for 
LBANN on Lassen as we had difficulties building the framework on ThetaGPU. 
Likewise, we only show AxoNN results on Lassen due to jobs not finishing on 
ThetaGPU.



\subsection{Evaluation Metrics}

For our analysis we use metrics that matter the most to a deep 
learning researcher - epoch execution times, statistical efficiency, and GPU 
memory consumption. Statistically efficient training algorithms or frameworks 
require less number of epochs to reach a certain target accuracy on the validation 
data. When comparing parallel DL frameworks it is absolutely imperative to 
compare both the epoch execution times and statistical efficiency of the 
training runs. We have discussed the tradeoffs that parallel DL algorithms 
incur between these two metrics in Section \ref{sec:survey}. 

We profile epoch execution times on 1, 2, 4, 8, 16, 32 and 64 GPUs on Lassen and 
ThetaGPU. While profiling the statistical efficiency 
for a particular framework, we use the GPU count where it has the minimum epoch 
execution times. For gathering memory utilization data we use 1, 2, 4, 8, 16, 32 and 
64 GPUs on ThetaGPU. Table \ref{tab:datasets} and Table \ref{tab:system-info} gives an overview of the neural 
networks and machines we used for evaluating these metrics.

\revision{
    To measure the statistical efficiency we record the accuracy and loss for the vision 
    tasks and perplexity for the language tasks.
    Loss is the output of the loss function used for training. Its magnitude depends on 
    its definition, but the training loss should decrease towards zero as the model 
    improves in predictive capacity.
    Accuracy measures the ratio of samples accurately predicted to total samples. We use
    the validation accuracy, which is calculated based on samples exclusive to the training set.
    Perplexity is commonly used in NLP to measure how well a model predicts for a certain corpus based on the cross-entropy of the model. It is defined as the exponential of the cross entropy loss on the dataset. 
}

%% file: results.tex
In this section we present and discuss the results from our experiments on epoch execution times, statistical efficiency, and memory utilization.

\begin{figure}[h]
  \includegraphics[width=0.45\textwidth]{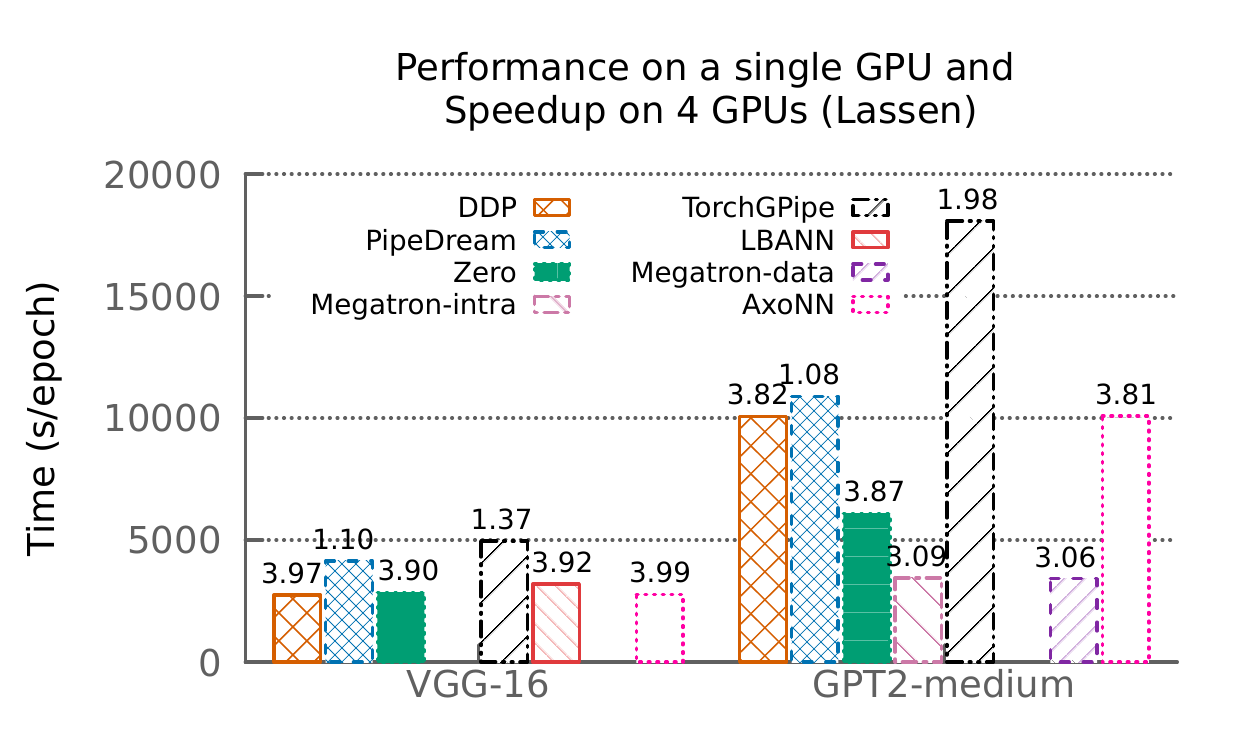}
  \caption{Comparison of single GPU performance and 4 GPU speedup on Lassen
          for VGG-16 and GPT2-medium. The labels list the speedup of each framework
          relative to their own 1 GPU performance.}
  \label{fig:single-gpu-perf}
\end{figure}

\subsection{Execution Time Comparison}
\label{sec:perf}

We first look at the baseline performance of each framework.
Figure~\ref{fig:single-gpu-perf} presents the sequential single GPU execution 
times on the two neural networks on Lassen. 
In this test TorchGPipe performs the worst on both VGG-16 and GPT2-medium by up to 
1.8x and 5.2x, respectively.
We also observe Pipedream is the second slowest framework.
The single GPU performances differ significantly largely due to these two not
supporting mixed precision.
The difference is exacerbated for extremely compute intensive neural networks 
like the GPT2-medium. 

While Megatron, DDP, ZeRO and AxoNN employ mixed precision, Megatron is 
considerably faster as it uses its own optimized implementation of the 
transformer encoder layer and Adam optimizer. Figure~\ref{fig:single-gpu-perf} exemplifies this, 
where we observe a 2x speedup on a single GPU over the native PyTorch kernel used by DDP and ZeRO. The PyTorch implementation performs worse due to its handling of 
the computationally intensive final softmax layer in GPT2-medium.
While DDP and AxoNN compute this layer in full precision, ZeRO's mixed 
precision strategy computes this layer in half precision,, 
leading to the difference in performance between the two. 

Out of all the frameworks TorchGPipe has the worst single GPU performance. 
This is because micro-batching provides no performance benefits as operations 
of different microbatches are serialized on a single GPU. It however does save 
memory used for stashing activations during the forward pass. We discuss this 
in Section \ref{sec:mem-use}.

\begin{figure}[h]
  \includegraphics[width=0.45\textwidth]{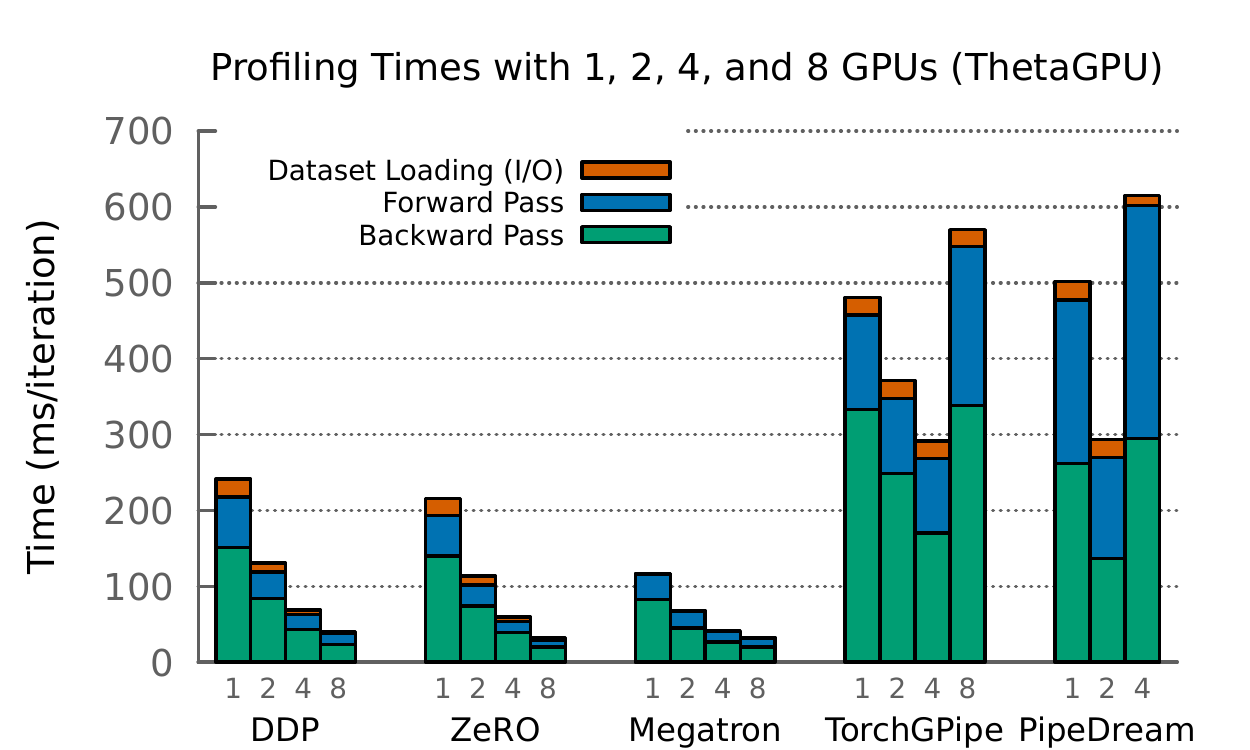}
  \caption{Breakdown of time spent in training on 1, 2, 4, and 8 GPUs of ThetaGPU for GPT2-medium. 
          We use NVIDIA's NVTX SDK for annotating events and Nsight Systems for instrumentation.
          Megatron refers to Megatron-intra.}
  \label{fig:theta-profiles}
\end{figure}

\begin{figure*}[t]
    \centering
      \includegraphics[width=0.45\textwidth]{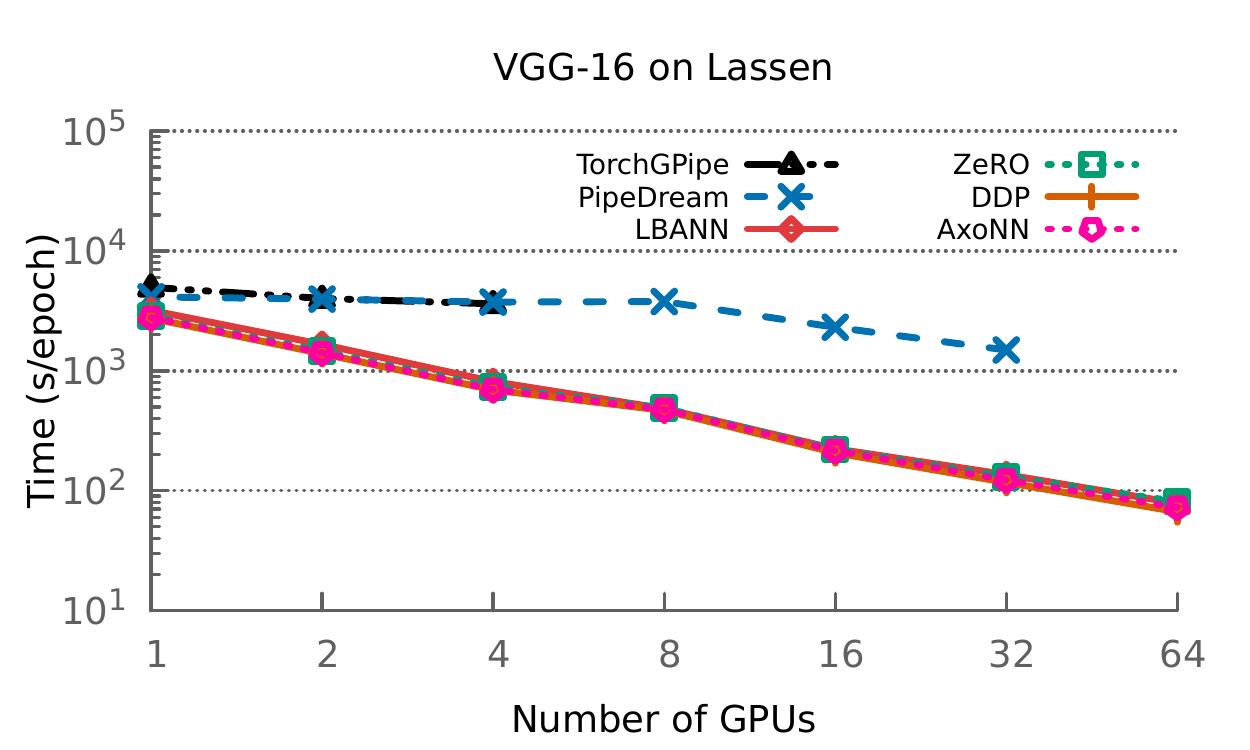}
      \includegraphics[width=0.45\textwidth]{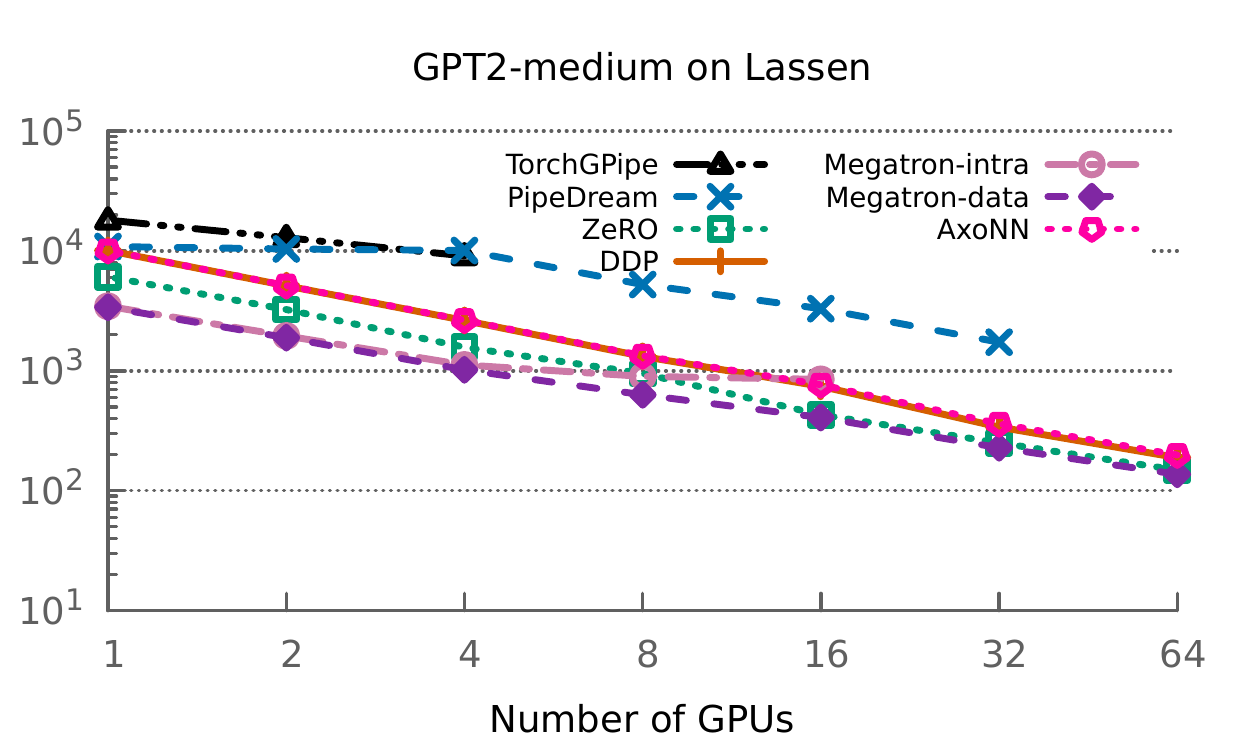}
    \caption{Performance results on Lassen for VGG-16 and GPT2-medium.}
    \label{fig:perf-lassen}
\end{figure*}

\begin{figure*}[t]
    \centering
      \includegraphics[width=0.45\textwidth]{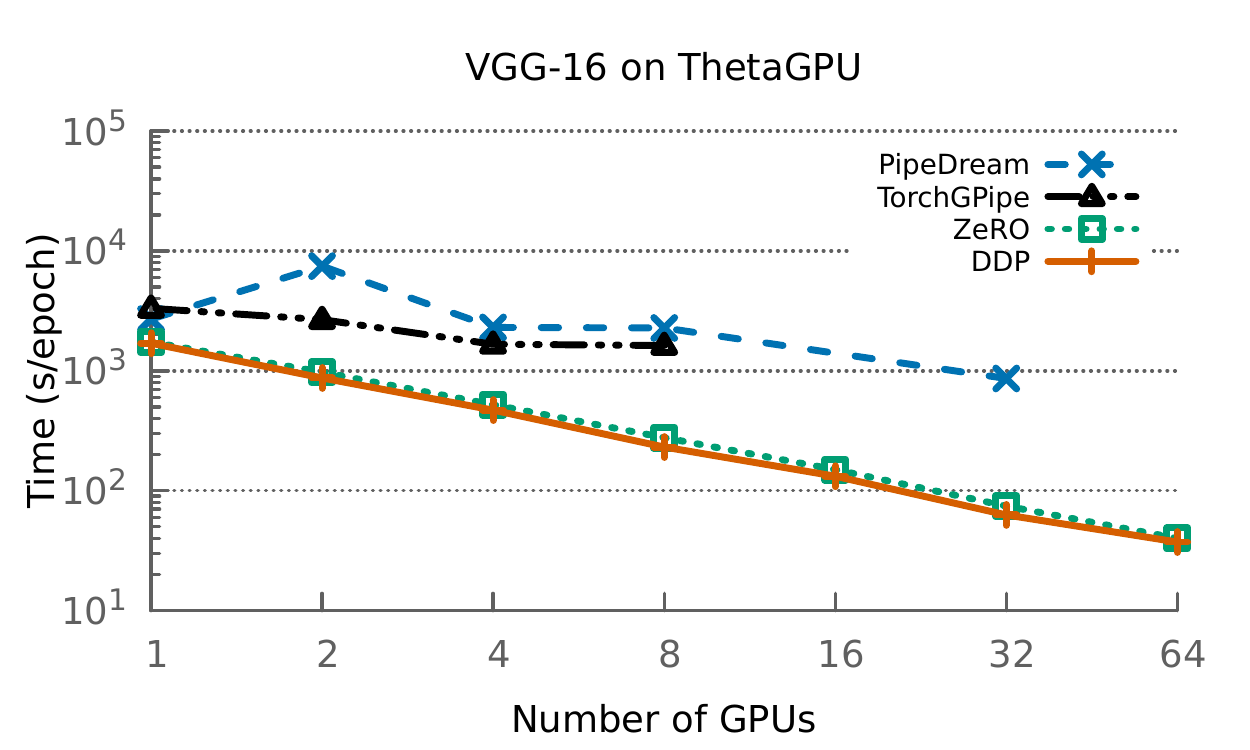}
      \includegraphics[width=0.45\textwidth]{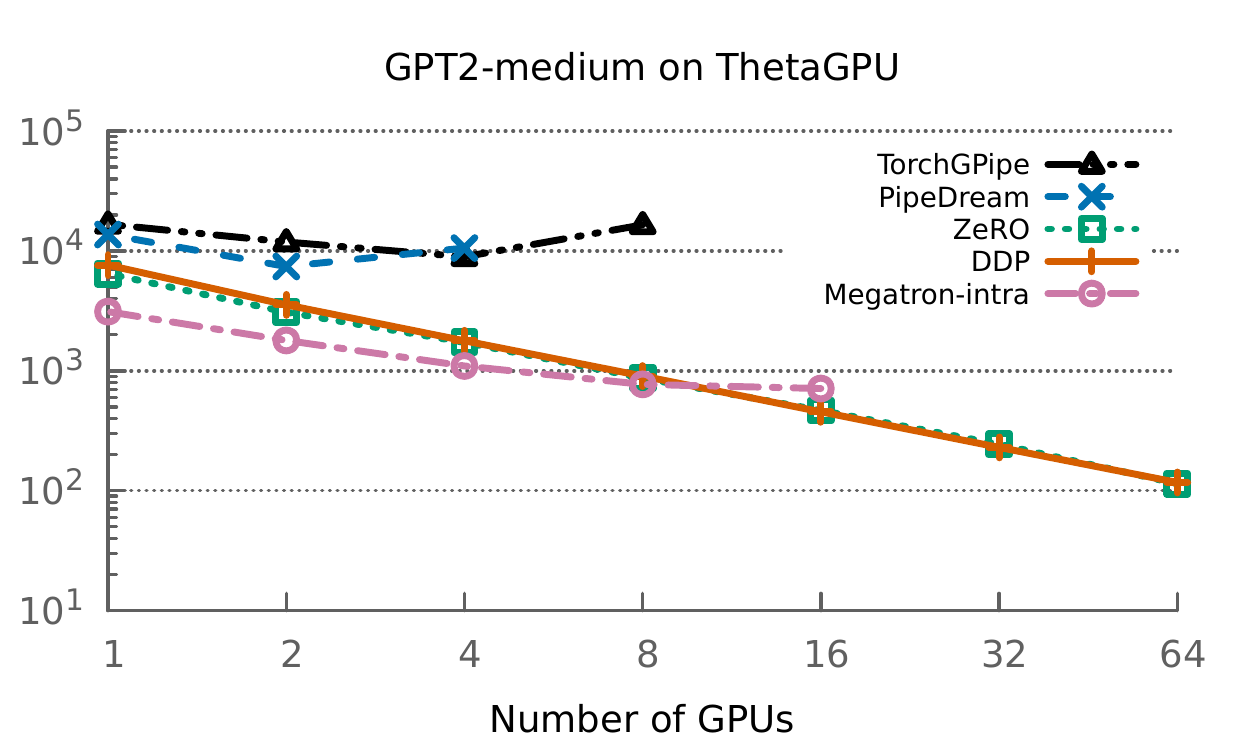}
    \caption{Performance results on ThetaGPU for VGG-16 and GPT2-medium.}
    \label{fig:perf-theta}
\end{figure*}

Figure~\ref{fig:theta-profiles} shows the time spent by each framework in the 
forward pass, backward pass, and I/O for GPT2-medium on ThetaGPU. We observe a 
marked improvement in Megatron's I/O performance due to 
its custom data loaders (see Section \ref{sec:choice-of-frameworks}), however,
these are a negligible part of the overall time per iteration. Across all frameworks, we see that the backward pass is more computationally intensive than the forward pass. This is because for each layer we not only compute the gradients for its parameters but also for its input activations which need to be backpropagated to previous layers.

Single GPU profiles in the figure also highlight the difference in the 
absolute computation time of the forward and backward passes for these frameworks. 
It further supports our above explanation for the differences in sequential
performance in Figure~\ref{fig:single-gpu-perf}.

Figures \ref{fig:perf-lassen} and \ref{fig:perf-theta}
detail the results from the performance tests on each machine. 
We present number of seconds per epoch for each neural network as the GPU 
count increases from 1 to 64.



Across both machines and neural networks we observe two separate trends amongst
the frameworks. First, DDP, ZeRO, LBANN, AxoNN and Megatron-data all perform 
similarly with only constant deviations from each other. Second, PipeDream and 
TorchGPipe are slower, more erratic, and scale worse than the others. Third, 
Megatron-intra's speedup seems to plateau when we try to scale it across 
multiple nodes.

Within this first trend we observe that ZeRO's performance trends the same as 
DDP and AxoNN with only 10-15\% difference in absolute run time. These 
variations can be attributed to the different mixed precision implementations 
and ZeRO's memory optimizations. As noted previously in Section \ref{dp_mem}, 
ZeRO reduces the per GPU memory footprint of data parallelism at the expense of 
added communication. However, we see that this communication overhead scales the
same as standard DDP.

It is immediately apparent that these data parallel approaches strongly 
outperform the other frameworks in scaling. 
This is notably due to the embarrassingly parallel workload in data parallelism 
when the entire model fits within GPU memory. 
We also see an expected slight reduction in speedup on Lassen and ThetaGPU 
(shown in Figure \ref{fig:single-gpu-perf}) for data parallelism as the number 
of GPUs surpassed that of a single node. This happens as the all-reduce 
communication now occurs outside the fast intra-node NVLink and has to use the 
system network.
This is a negligible issue due to how much better the data parallel algorithms 
scale.

Due to the lack of mixed precision support, PipeDream and TorchGPipe have the 
largest epoch execution times at all GPU counts across all machines. 
PipeDream seems to scale erratically relative to its own single 
GPU execution. The poor scaling can be attributed to two factors. Firstly, 
PipeDream uses the relatively slow Gloo library as its communication backend. 
Secondly, erratic scaling is usually a sign of load imbalance. Our 
experiments show that their communication-aware load balancing algorithm does 
not perform satisfactorily in practice.


Along with these two major trends we also observe that Megatron-intra plateaus
once it runs on multiple nodes. For larger GPU counts it scales worse than 
DDP, ZeRO and AxoNN.
We observed that the communication overhead of Megatron-intra increases rapidly with 
increasing number of GPUs, ultimately reaching 52.5\% of the total execution 
time on 16 GPUs. Based on our observations we recommend that 
researchers who wish to train large transformer models on 
language modeling task use Megatron-intra for their single GPU sequential 
implementations. If the model surpasses the memory capacity of a single GPU, we 
recommend employing Megatron's intra-layer parallelism to fit the model inside the 
GPUs of a single node. Scaling to large GPU counts should be done by 
integrating Megatron's intra-layer parallelism with data parallelism. 



\subsection{Statistical Efficiency}

\begin{figure*}[h]
  \centering
  \includegraphics[width=0.49\textwidth]{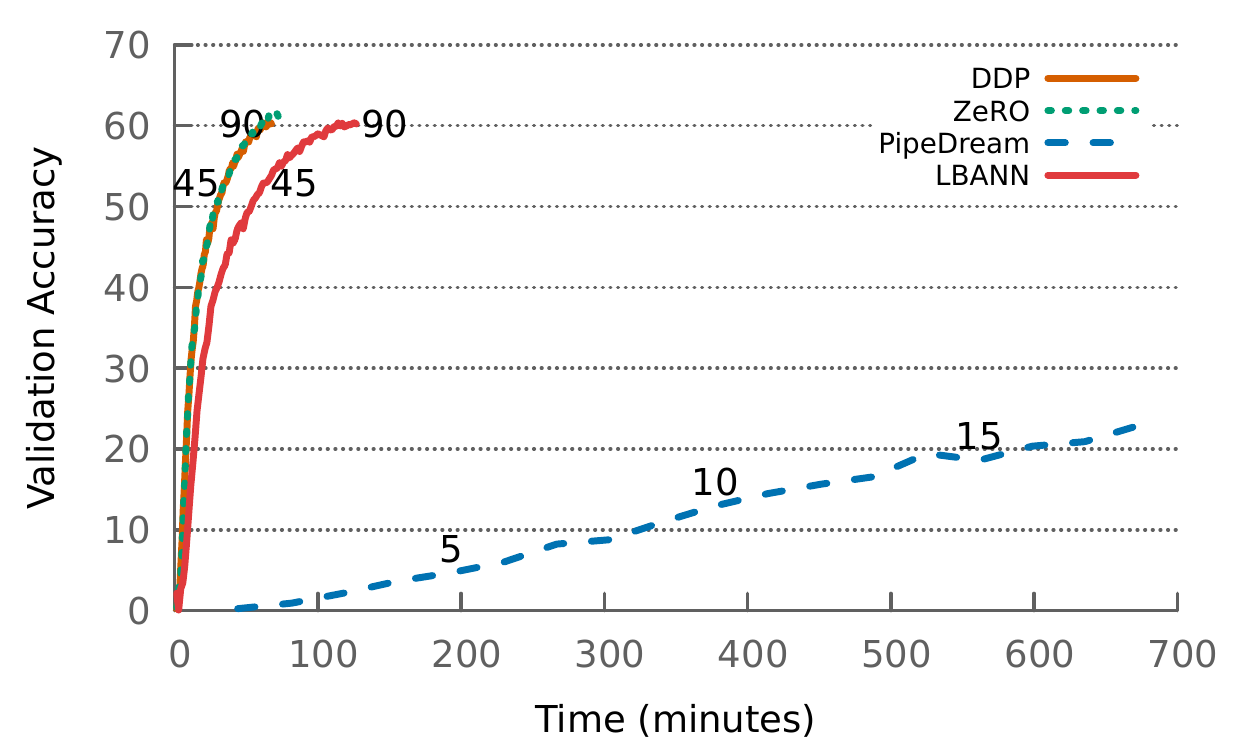}
  \includegraphics[width=0.49\textwidth]{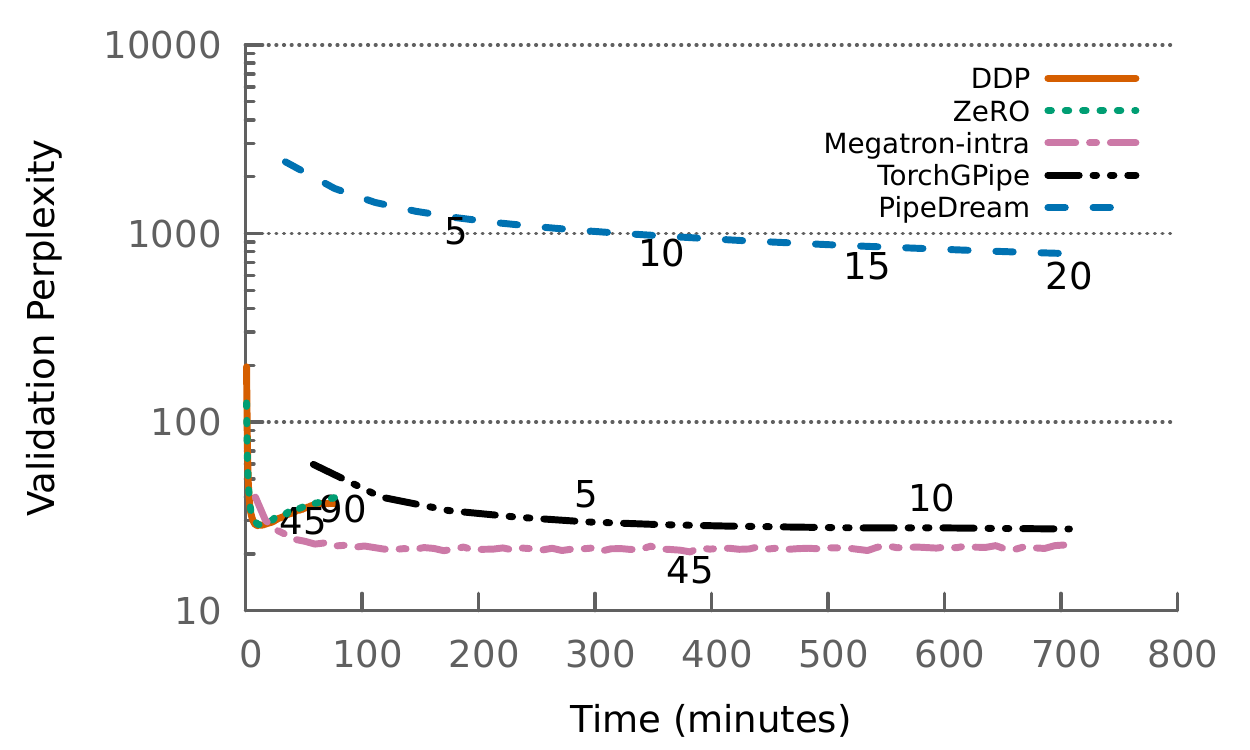}
  \caption{Validation performance by time for training VGG-16 and GPT2-medium on ThetaGPU.
           Epoch numbers are shown in labels.}
  \label{fig:eff-all-time}
\end{figure*}

Figure \ref{fig:eff-all-time} illustrates the results of our statistical 
efficiency experiments. 
Following standard practice we measure the validation accuracy and perplexity 
at each epoch for the image classification and language modeling tasks respectively.
We report the epoch number as well as the total training time. 
On observing the performance of PipeDream on both the tasks it is apparent 
that weight staleness is a huge roadblock in the path of algorithms that seek 
to implement pipelining without flushing. 
PipeDream's proposed weight stashing approach does not mitigate this problem 
satisfactorily. 
ZeRO, DDP and LBANN exhibit near identical validation curves. 
The slight variations in the validation curves are likely due to differences 
in the mixed precision implementations in these frameworks. 
TorchGPipe and Megatron-intra exhibit greater statistical efficiencies than the 
data parallel frameworks on the language modeling task. 
We attribute the fast convergence of these frameworks due to their training 
runs being carried out on a small GPU count. 
The data parallel frameworks being trained at 64 GPUs take a slight hit in 
their convergence speeds due to the problem of increase effective 
mini-batch sizes that we highlighted in Section~\ref{lars}.

Figure \ref{fig:eff-all-time} further details how the accuracies and 
perplexities behave over time rather than epoch.
PipeDream is much slower to accuracies than the other frameworks. 
Such a figure presents a combined picture of the statistical efficiency and 
epoch execution times of a framework. 
We argue that plotting validation metrics against epoch times is the best way 
to evaluate the performance of any distributed deep learning framework. 
It also clearly demonstrates the superiority of data parallelism over other 
classes of parallel deep learning algorithms.



\subsection{Memory Utilization}\label{sec:mem-use}

\begin{figure}
  \includegraphics[width=0.45\textwidth]{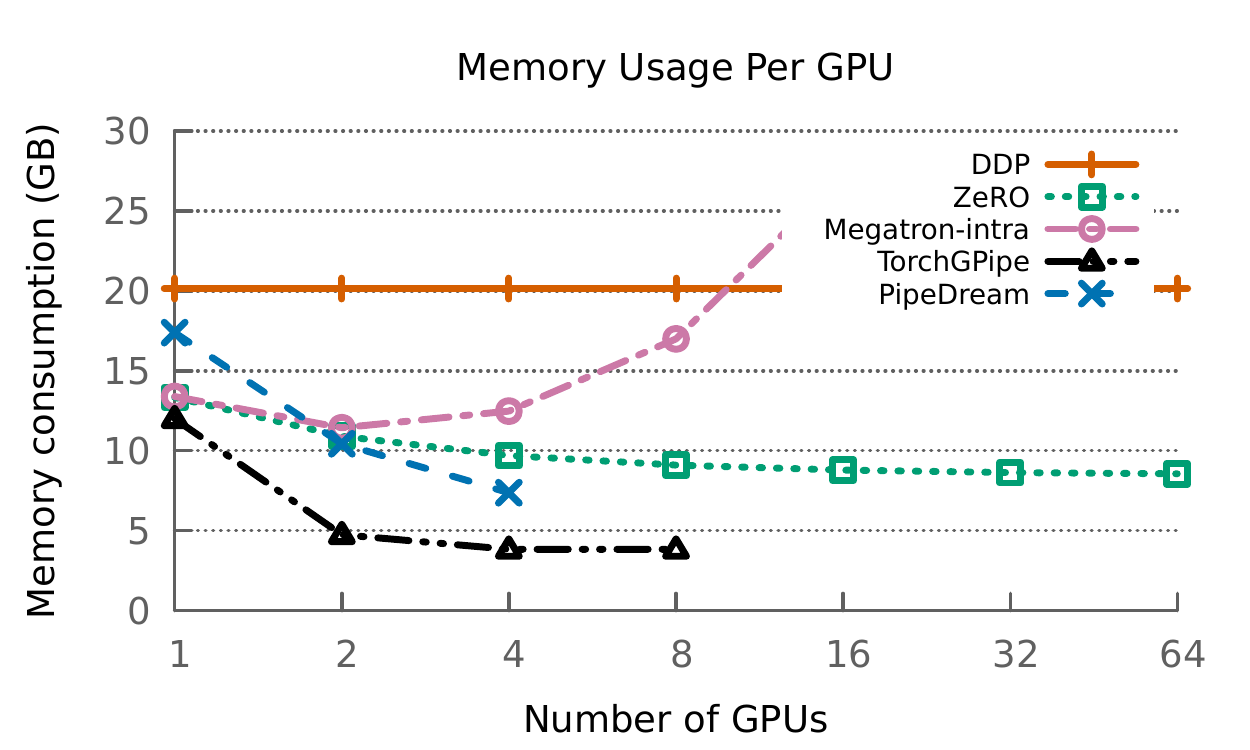}
  \caption{Memory consumption by different frameworks on ThetaGPU for GPT2-medium.}
  \label{fig:memory-usage}
\end{figure}

Figure~\ref{fig:memory-usage} details the per GPU memory usage of each 
framework during the training tasks. ZeRO, while having similar performance and 
scaling to DDP, had between 42\% and 66\% of the memory footprint.
We also see this improving as more GPUs are added similar to the layer parallel 
runs, while DDP remains fixed as it simply duplicates the models 
across GPUs. 

The pipelining implementations both experienced over 2x better memory usage 
with more resources. More of the models were able to be partitioned amongst the 
GPUs. However, the memory savings begin to plateau as more GPUs are added since 
increase in the activation memory due to increasing batch sizes balances out 
the decrease in parameter memory.

The U-shaped per GPU memory curve of Megatron can be attributed to the inner 
workings of their intra-layer parallelism implementation. While the 
computation of a transformer layer is divided across multiple GPUs, the output 
of the last layer needs to be present in its entirety on every GPU. Since the 
per GPU mini-batch size is fixed the memory occupied by the input for any 
layer on each GPU increases linearly with an increase in GPU count. At lower 
GPU counts this increase is offset by the decrease in parameter memory due to 
the division of the layer computation across GPUs. After a while, however, the 
decrease is not enough to completely offset the increasing input activation 
memory.

%% file: conc.tex
The increasing size of contemporary neural network architectures has 
necessitated the development of efficient algorithms for parallelizing neural 
networks. The performance of parallel
training of neural networks is heavily dependent on the algorithm,
implementation, hyperparameters, and hardware used. In this paper we provide a
comprehensive survey of parallel deep learning frameworks that have
demonstrated scaling on parallel systems. We use two dataset-network
combinations to study various properties of parallel deep learning frameworks
such as scalability, memory requirements, and statistical efficiency as a
function of performance. 

Our benchmarking studies presents some interesting observations. When the
entire model can fit within a single GPU, it is best to use data parallel
approaches as they perform and scale well. In memory constrained environments, 
ZeRO~\cite{sc2020zero} can save us a decent amount of memory. Their memory optimizations only add 
substantial cost to the computation for non-transformer models. For saving more 
memory we recommend using intra or inter-layer parallelism to deploy a model 
across a few number of GPUs and then scale it in a hybrid fashion with data 
parallelism.